\documentclass[runningheads]{llncs}
\usepackage[T1]{fontenc}
\usepackage{booktabs}
\usepackage{tabularx}
\usepackage{multirow}
\usepackage{multirow}
\usepackage{amsmath, amssymb}
\usepackage{cite}
\usepackage[table]{xcolor}
\usepackage{hhline}
\usepackage{soul}
\usepackage{float}
\soulregister{\cite}7 
\soulregister{\citep}7 
\soulregister{\citet}7 
\soulregister{\ref}7 
\soulregister{\pageref}7 
\soulregister{\say}7 

\usepackage[colorlinks=true, citecolor=blue]{hyperref}
\hypersetup{
    linkcolor= blue, 
    urlcolor= blue, 
}
\usepackage{dirtytalk}

\usepackage{graphicx}

\usepackage{booktabs}
\usepackage{adjustbox}
\usepackage{amssymb}
\usepackage{todonotes}
\usepackage{multirow}
\usepackage{xcolor, colortbl}
\usepackage{hyperref}
\usepackage{marvosym}
\usepackage{float} 
\usepackage{marvosym} 
\newcolumntype{L}[1]{>{\raggedright\arraybackslash}p{#1}}

\definecolor{cgreen}{rgb}{0,0.7,0.8}
\definecolor{cred}{rgb}{0.968,0.545,0.321}
\definecolor{brightpink}{rgb}{1.0, 0.0, 0.5}

\newcommand{\gtr}[1]{{\scriptsize\color{cgreen}$\blacktriangle$ #1}}
\renewcommand{\thefootnote}{\fnsymbol{footnote}} 
\begin{document}

\title{Language Models Meet Anomaly Detection for Better Interpretability and Generalizability}
\titlerunning{Language Models Meet Anomaly Detection}

\author{Jun Li\inst{1,2} \and Su Hwan Kim\inst{4} \and  Philip Müller\inst{1} \and Lina Felsner\inst{1}  \and Daniel Rueckert\inst{1,2,4,5} \and Benedikt Wiestler\inst{1,4} \and Julia A.Schnabel*\inst{1,2,3,6}\textsuperscript{(\Letter)} \and Cosmin I. Bercea* 
 \inst{1,3}\textsuperscript{(\Letter)}} 

\authorrunning{Jun Li et al.}
\renewcommand{\thefootnote}{\fnsymbol{footnote}} 
\footnotetext[1]{Equal contribution.}

\institute{Technical University of Munich, Germany  \and
Munich Center for Machine Learning, Germany \and
Helmholtz AI and Helmholtz Munich, Germany \and
Klinikum Rechts der Isar, Munich, Germany \and
Imperial College London, UK \and
King’s College London, UK \\
\email{\{june.li,julia.schnabel,cosmin.bercea\}@tum.de}
}

\maketitle              

\begin{abstract}
This research explores the integration of language models and unsupervised anomaly detection in medical imaging, addressing two key questions: \textit{(1) Can language models enhance the interpretability of anomaly detection maps?} and \textit{(2) Can anomaly maps improve the generalizability of language models in open-set anomaly detection tasks?} To investigate these questions, we introduce a new dataset for multi-image visual question-answering on brain magnetic resonance images encompassing multiple conditions. We propose \emph{KQ-Former} (Knowledge Querying Transformer), which is designed to optimally align visual and textual information in limited-sample contexts.  
Our model achieves a 60.81\% accuracy on closed questions, covering disease classification and severity across 15 different classes. For open questions, \emph{KQ-Former} demonstrates a 70\% improvement over the baseline with a BLEU-4 score of 0.41, and achieves the highest entailment ratios (up to 71.9\%) and lowest contradiction ratios (down to 10.0\%) among various natural language inference models. Furthermore, integrating anomaly maps results in an 18\% accuracy increase in detecting open-set anomalies, thereby enhancing the language model's generalizability to previously unseen medical conditions. The code and dataset are available at:  \url{https://github.com/compai-lab/miccai-2024-junli?tab=readme-ov-file}.


\begin{figure*}[h]
\centerline{\includegraphics[width=0.85\textwidth]{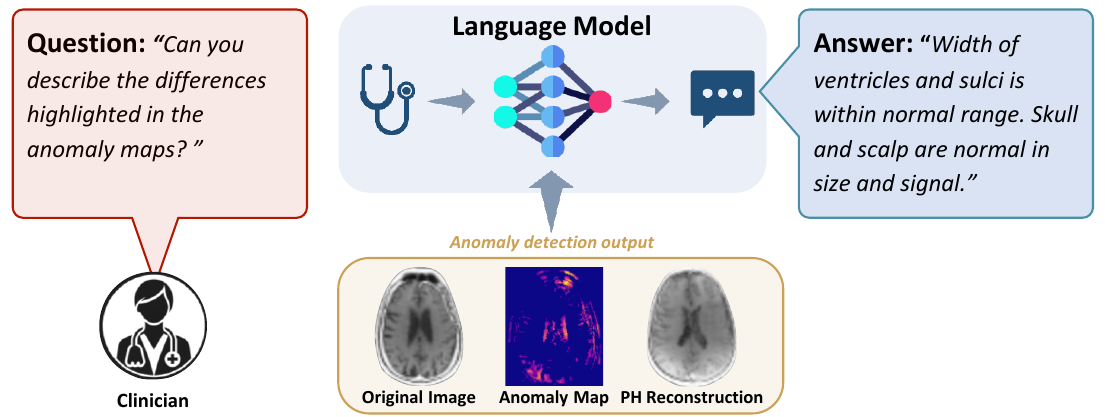}}
\caption{\label{fig1}Our framework is designed to process questions in conjunction with results from anomaly detection methods aiming to provide clinicians with clear, interpretable responses that render anomaly map analyses more intuitive and clinically actionable. }

\end{figure*}

\end{abstract}

\section{Introduction}
Unsupervised Anomaly Detection (UAD) plays a vital role in early disease diagnosis by identifying deviations from normal patterns. Common UAD methods in medical imaging utilize auto-encoders~\cite{chen2020unsupervised,zimmerer2019unsupervised}, generative adversarial networks~\cite{schlegl2019fanogan,akcay2019ganomaly}, or diffusion models~\cite{behrendt2023patched,bercea2023mask,wolleb2022diffusion} and are typically trained on data from healthy subjects. When applied to pathological inputs, they generate counterfactual "pseudo-healthy" (PH) images that normalize anomalous features to resemble healthy tissues~\cite{bercea2023reversing}. By comparing the pathological inputs with the generated PH images, anomaly maps can be derived, highlighting regions of interest for clinicians. However, the interpretability of UAD findings is inherently limited due to the unsupervised nature of these methods. Clinicians often lack explicit explanations of the detected anomalies, which can hinder effective decision-making.

To provide interpretable text descriptions for clinicians, we integrate language models with UAD as shown in Figure~\ref{fig1}. Recent advancements in language models have achieved human-like performance in tasks such as question answering, summarizing, reasoning, and knowledge retrieval~\cite{openai_chatgpt,touvron2023llama,chowdhery2023palm}. Notably, these models have demonstrated the capability to pass the United States medical licensing examination~\cite{kung2023performance}, showcasing their potential in the medical field~\cite{tu2024towards,singhal2023large,luo2022biogpt}.

However, integrating language models with UAD shifts the task from a typical single-image analysis~\cite{bai2023cat,liu2021contrastive,van2023open} to a more complex multi-image visual question answering (VQA) challenge. While recent studies have explored generating radiology reports from frontal and lateral views of X-rays~\cite{wu2022deltanet,li2022self,Tanida_2023_CVPR}, the broader application of multi-image VQA remains largely unexplored. To address this gap, we propose a framework for multi-image VQA in UAD, analyzing various feature fusion strategies to effectively combine original images, anomaly maps, and PH reconstructions. Furthermore, adapting language models for multi-modal tasks introduces additional challenges due to the scarcity and high costs associated with large annotated medical datasets required for fine-tuning~\cite{li2023masked,zhang2023pmc,nguyen2019overcoming,li2023blip}. To tackle these challenges, we introduce the \emph{KQ-Former}, a novel module designed to improve the alignment between visual and textual features, even in settings with limited data availability.
In this work, we propose, to the best of our knowledge, the first multi-image question answering application for unsupervised anomaly detection (VQA-UAD). Our main contributions are as follows:

\begin{itemize}
    \item We have developed a specialized multi-image VQA dataset for UAD, featuring brain Magnetic Resonance Imaging scans. This dataset is meticulously annotated by medical experts and covers a wide range of medical conditions.
    \item We introduce a model-agnostic baseline tailored for multi-image VQA-UAD, alongside a comprehensive analysis of various image fusion strategies.
    \item We introduce the \emph{KQ-Former}, an innovative module designed to enhance the extraction of knowledge-related visual features, thereby improving the alignment between visual and textual information.
    \item Our experimental results demonstrate that language models not only render anomaly maps interpretable but can also leverage these anomaly maps to bolster the accuracy of responses in VQA. This proved particularly effective in scenarios involving previously unseen anomalies.
\end{itemize}
\section{Methods}
Figure \ref{fig2} shows our VQA-UAD framework, which leverages multiple imaging modalities and language models to enhance diagnostic accuracy. The goal of VQA-UAD is to generate precise answers $(A_i)$ from a set of three images—original image $(I_i^o)$, anomaly map $(I_i^a)$, and PH reconstruction $(I_i^r)$—and a question $(Q_i)$. Section \ref{subsec:baseline} introduces our baseline for multi-image VQA, setting the foundation for this application. Section \ref{subsec:framework} introduces the novel \emph{KQ-Former}, designed to enhance both UAD and VQA through improved visual-textual alignment.
\begin{figure*}[t]
\centerline{\includegraphics[width=\textwidth]{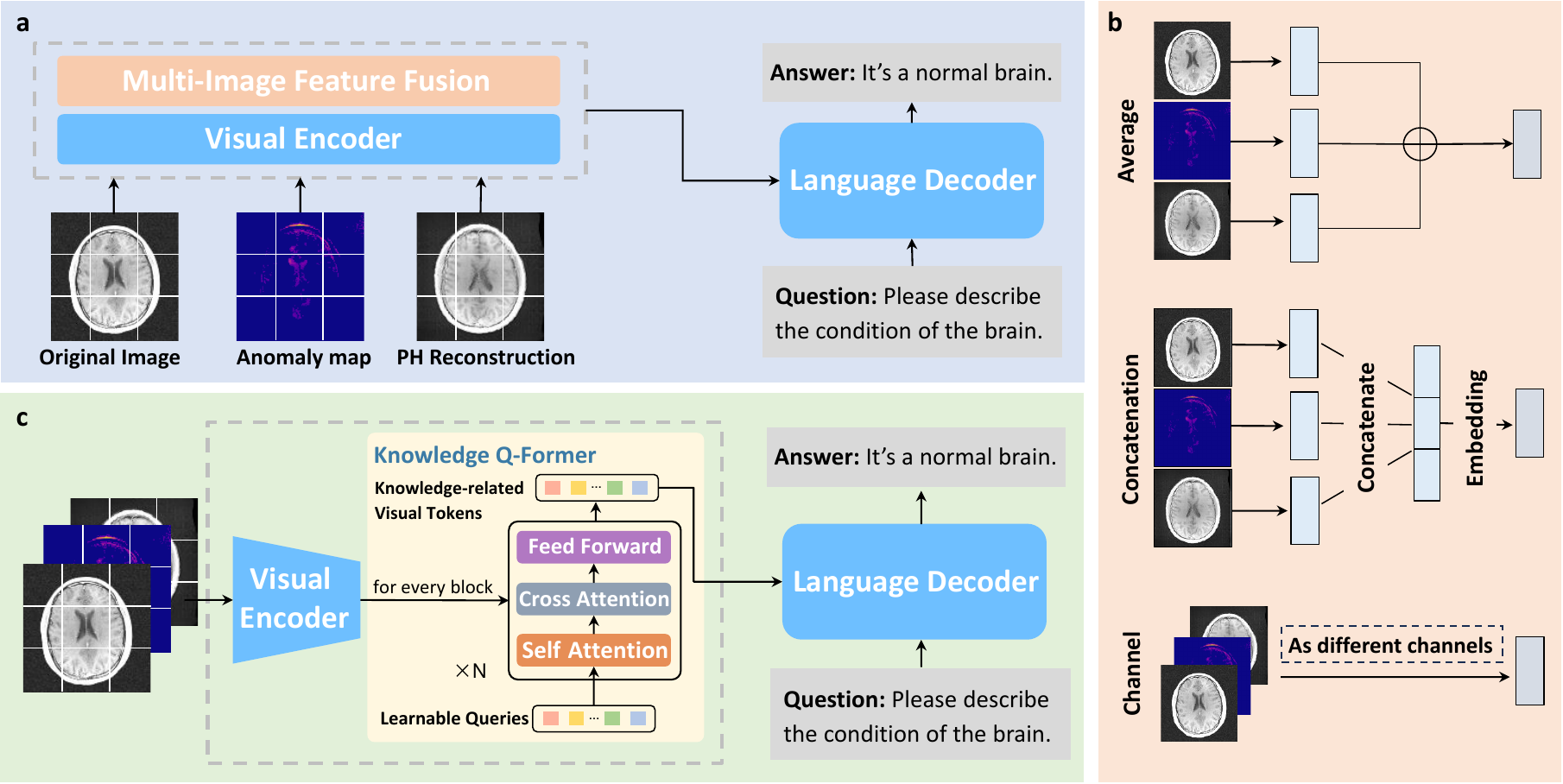}}
\caption{An overview of our novel framework for VQA-UAD: (a) the multi-image VQA baseline; (b) multi-image feature fusion strategies; (c) the \emph{KQ-former} module.}
\label{fig2}
\end{figure*}
\subsection{Multi-Image VQA Baseline}
\label{subsec:baseline}
Figure \ref{fig2}a provides an overview of our multi-image VQA baseline, which incorporates a visual encoder and a language decoder. Figure \ref{fig2}b illustrates three feature fusion methods within our module.\\

\noindent\textbf{The visual encoder} processes the image triple $\mathcal{I} = (I_i^o, I_i^a, I_i^r)$, where $I_i$ is a tensor in $\mathbb{R}^{H \times W \times C}$ representing height, width, and channels, respectively. It transforms these images into visual embeddings $\mathcal{V} = (V_i^o, V_i^a, V_i^r)$ through the operation $\mathcal{V} = \mathcal{F}_v(\mathcal{I})$, where each $V_i$ is an array in $\mathbb{R}^{n \times d}$, with $n$ indicating the number of patches, and $d$ the dimension of embeddings. Here, we implement the encoder based on two different backbones: ViT-B/16 \cite{dosovitskiy2020image} and ResNet50 \cite{he2016deep}. \\

\noindent\textbf{The different fusion strategies} are depicted in Figure~\ref{fig2}b. The first strategy averages the image features by computing $V_i' = \frac{1}{3} \sum_{j \in \{o, a, r\}} V_i^j$. The second strategy concatenates the visual features into $C_i = [V_i^o; V_i^a; V_i^r]$, where $C_i$ resides in $\mathbb{R}^{3n\times d}$. Subsequently, a trainable projection model $\Phi(\cdot)$ is employed to reduce the dimension of $C_i$ into $V_i^{'} = \Phi(C_i)$, where $\Phi(\cdot)$ consists of a two-layer  multilayer perception. The final fusion strategy converts each component of the image triple $\mathcal{I}$ into single-channel grayscale images. These are then concatenated channel-wise to form a combined three-channel image $\hat I_i = [I_i^o, I_i^a, I_i^r] \in \mathbb{R}^{H \times W \times 3}$. This transformation simplifies the multi-image VQA challenge into a single-image format, with the final integrated visual features expressed as $V_i' = \mathcal{F}_v(\hat I_i)$.\\

\noindent\textbf{Language Decoder.} Unlike existing methods that primarily treat VQA as a classification task~\cite{nguyen2019overcoming,liu2021contrastive}, our framework approaches it as a natural language generation challenge, drawing inspiration from recent advances~\cite{wu2022deltanet,Tanida_2023_CVPR, li2022self}. Our language decoder, here GPT-2 small~\cite{radford2019language}, processes a question $Q_i$ and the corresponding merged image features $V_i'$ to generate the answer $A_i$, producing tokens sequentially. At each decoding step $t$, it calculates a probability distribution $p_\theta(A_i^t)$ over the vocabulary. Consequently, the full answer distribution for $A_i$ with length $T$ is defined as $p_\theta(A_i) = \prod_{t=1}^T p_\theta(A_i^t | Q_i, V_i', A_i^1, \ldots, A_i^{t-1})$. During training, the goal is to minimize the negative log-likelihood for $N$ samples:
\begin{equation}
    \mathcal{L}(\theta^*) = \arg\min_\theta \sum_{i=1}^N -\log p_\theta(A_i | Q_i, V_i').
\end{equation}

\subsection{Knowledge Q-Former}
\label{subsec:framework}

We introduce the \emph{KQ-Former} as an enhancement to our baseline multi-image VQA, specifically addressing the challenge of effectively aligning visual and textual features in contexts where datasets are typically limited. Figure \ref{fig2}c illustrates the design and integration of the \emph{KQ-Former} within our VQA framework. The novel aspect of the \emph{KQ-Former} lies in its ability to leverage knowledge embeddings that integrate pre-trained medical information, enriching its capability to understand and interpret complex medical images. 

Formally, the \emph{KQ-Former} operates by taking an input of learnable queries $L_i \in \mathbb{R}^{32 \times 768}$ and visual features $V_i$ and processes these through a dynamic cross-attention mechanism. This interaction enables the \emph{KQ-Former} to dynamically merge the embedded medical knowledge with the visual information, resulting in enhanced visual tokens $K_i \in \mathbb{R}^{32 \times 768}$ that carry detailed visual data alongside relevant medical insights. These tokens are then aggregated following the strategies outlined in Section \ref{subsec:baseline} and fed directly into the language decoder.\\

\noindent\textbf{Network Architecture.} 
The architecture of the \emph{KQ-Former} is transformer-based, inspired by the Q-Former design \cite{li2023blip,vaswani2017attention} but modified to consolidate image and text processing into a single transformer unit. This simplification is crucial in medical applications, where data sets are often limited to a few hundred samples. Additionally, the KQ-Former is initialized with BioBERT \cite{lee2020biobert}, enhancing its ability to incorporate deep medical knowledge.

\section{Experiments}

\noindent\textbf{Dataset.}

We retrieved 440 T1-weighted MRI 2D mid-axial brain images from the fastMRI dataset \cite{knoll2020fastmri}, including 253 healthy and 187 unhealthy samples, featuring 13 distinct types of anomalies. For our main experiment, we focused on seven types, while the remaining six types were used to test open-set anomaly detection capabilities (refer to supplementary material for category distribution). We generated the anomaly maps and PH reconstructions using the publicly available method in~\cite{bercea2023generalizing}. Nevertheless, our framework is complementary to UAD research and can benefit from advances in this field. 

\begin{figure*}[t]
\centerline{\includegraphics[width=\textwidth]{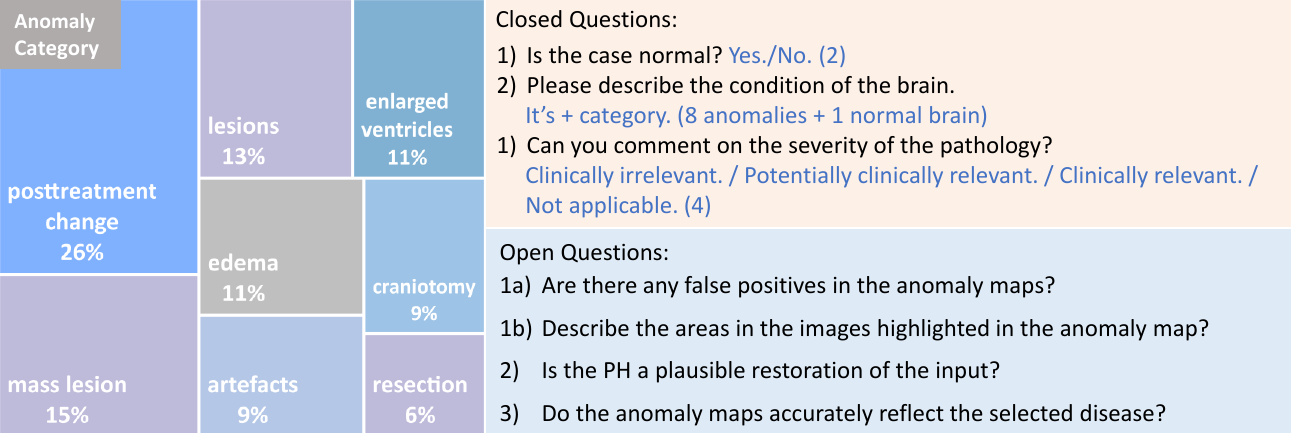}}

\caption{Left: Distribution of anomaly categories. Right: Definitions of closed and open questions. For the closed questions, the blue text indicates the answer type, with the count of each type in parentheses. For more details, please refer to the supplementary material. Some questions are simplified here due to space constraints.}
\label{fig3}
\end{figure*}

We created and released VQA labels to facilitate further research. The dataset, annotated by two senior neuroradiologists with both closed and open question types as shown in Figure \ref{fig3}, is organized into question-answer pairs. We divide them patient-wise into training, validation, and test sets in a 7:1:2 ratio, containing 1078, 154, and 308 samples respectively, ensuring a diverse representation of disease and question types across all sets without overlap.\\

\noindent\textbf{Evaluation Metrics.}
We evaluated the performance on the closed questions using Accuracy (ACC) and F1 scores. For the open questions, we employed two types of metrics. Firstly, we used standard language evaluation metrics such as BLEU scores \cite{papineni2002bleu}, ROUGE-L \cite{lin2004rouge}, and CIDEr \cite{vedantam2015cider} to assess the similarity between the predicted answers and the ground truths. However, since these metrics primarily measure similarity without confirming factual accuracy, we supplemented them with a second type of evaluation. We utilized four Natural Language Inference (NLI) models—BART \cite{lewis-etal-2020-bart}, DEBERTA \cite{he2022debertav3}, mDeBERTa \cite{sileo2023tasksource}, and ROBERTA \cite{liu2019roberta}—to determine the logical relationship between the predicted answers and the ground truths. The NLI model categorizes whether the given predicted sentence and the ground truth answer logically imply (entailment) or oppose (contradiction) each other, or are indeterminate (neutral) to each other. \\

\noindent\textbf{Experimental Setup.} Our experiments focus on two main areas. The first evaluates how well our proposed methods explain anomaly maps. The second investigates whether anomaly maps can enhance the generalizability of language models in real-world clinical scenarios, which include predominantly healthy data and some previously unseen anomalies. We trained the models on a single NVIDIA RTX A6000 for $40$ epochs, using early stopping with patience of $10$. We utilized the AdamW optimizer \cite{loshchilov2017decoupled} with a learning rate of $1.5e^{-5}$ and a weight decay of $0.05$. We used beam search with a width of $5$ during the generation phase. 

\section{Results}

\begin{figure*}[tb!]
\includegraphics[width=\textwidth]{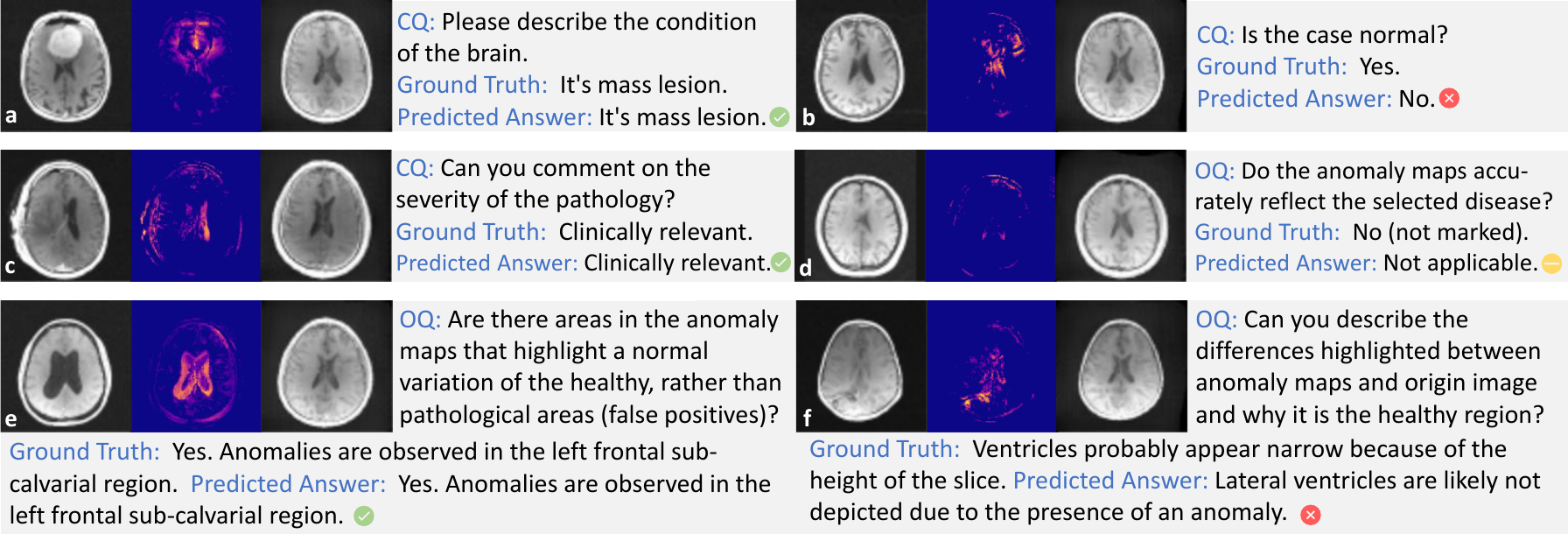}
\caption{Visualization examples of the \emph{KQ-Former} module with concatenation strategy. Each example includes from left to right: the original image, anomaly map, and PH reconstruction. CQ and OQ represent closed and open questions, respectively.}
\label{fig4}
\end{figure*}

\subsection{Language Models Enhance the Explainability of Anomaly Maps} 
\begin{table}[!t]
\caption{Performance of the proposed Multi-Image baseline (MI) and \emph{KQ-Former} (KQF). We experiment with different backbones: ViT and ResNet, and using different feature fusion strategies. B1 to B4 denote BLEU-1 to BLEU-4, while RL and Cr denote ROUGE-L and CIDEr. The best two performances are shown in \textbf{bold}.} 
\centering
\setlength{\tabcolsep}{4pt}
\begin{adjustbox}{width=0.9\linewidth,center} 
\begin{tabular}{l l | l | c c | c c c c c c}
\toprule
  \multicolumn{1}{l}{\multirow{2}{*}{}}&\multicolumn{1}{c|}{\multirow{2}{*}{Methods}} & \multicolumn{1}{l|}{\multirow{2}{*}{Fusion}} & \multicolumn{2}{c|}{Closed} & \multicolumn{6}{c}{Open} \\ 
\multicolumn{1}{l}{} &\multicolumn{1}{c|}{} &\multicolumn{1}{c|}{} & ACC$\uparrow$ & F1$\uparrow$ & B1$\uparrow$ & B2$\uparrow$ & B3$\uparrow$ & B4$\uparrow$ & RL$\uparrow$ & Cr$\uparrow$ \\ 
\midrule 
\multirow{5}{*}{\rotatebox{90}{ViT}}& \multirow{3}{*}{MI-baseline}  &average & 56.76 & 49.30 & 0.49 & 0.38 & 0.32 & 0.27 & 0.62 & 1.94 \\
          &{}&concat   &57.43 & 50.14 & 0.45 & 0.35 & 0.29 & 0.24 & 0.61 & 1.79 \\
          &{}& channel & 53.38 & 48.33 & 0.44 & 0.34 & 0.28 & 0.24 & 0.58 & 1.77 \\\cmidrule{3-11}
  &\multirow{2}{*}{KQF (ours)} & concat  & \textbf{60.14} & \textbf{56.92} & \textbf{0.55} & \textbf{0.48} & \textbf{0.44} & \textbf{0.41} & \textbf{0.67} & \textbf{2.84} \\
         &{}& channel  & \textbf{60.81} & \textbf{55.93} & \textbf{0.51} & \textbf{0.43} & \textbf{0.38} & \textbf{0.34} & \textbf{0.65} & \textbf{2.50}\\

\midrule
\midrule
\multirow{5}{*}{\rotatebox{90}{ResNet}}& \multirow{3}{*}{MI-baseline}  &average &36.49 & 38.89 & 0.44 & 0.35 & 0.29 & 0.24 & 0.57 & 1.82 \\
           &{}&concat   & 40.54 &\textbf{43.89} &0.35 &0.26 &0.21 & 0.18 & 0.56 & 1.48 \\
           &{}& channel & 36.49 & 38.89 & 0.38 & 0.29 & 0.24 & 0.20 & 0.54 & 1.69 \\\cmidrule{3-11}
            
           &\multirow{2}{*}{KQF (ours)} & concat  &\textbf{54.05} & \textbf{47.27} & \textbf{0.47} & \textbf{0.37} & \textbf{0.30} & \textbf{0.25} & \textbf{0.60} & \textbf{1.97} \\
           &{}& channel  & \textbf{47.97} & 41.54 & \textbf{0.45} & \textbf{0.36} & \textbf{0.31} & \textbf{0.27} & \textbf{0.58} & \textbf{1.90} \\
\bottomrule
\end{tabular}
\label{table1}
\end{adjustbox}
\end{table}

\begin{figure*}[!h]
\centerline{\includegraphics[width=0.9\textwidth]{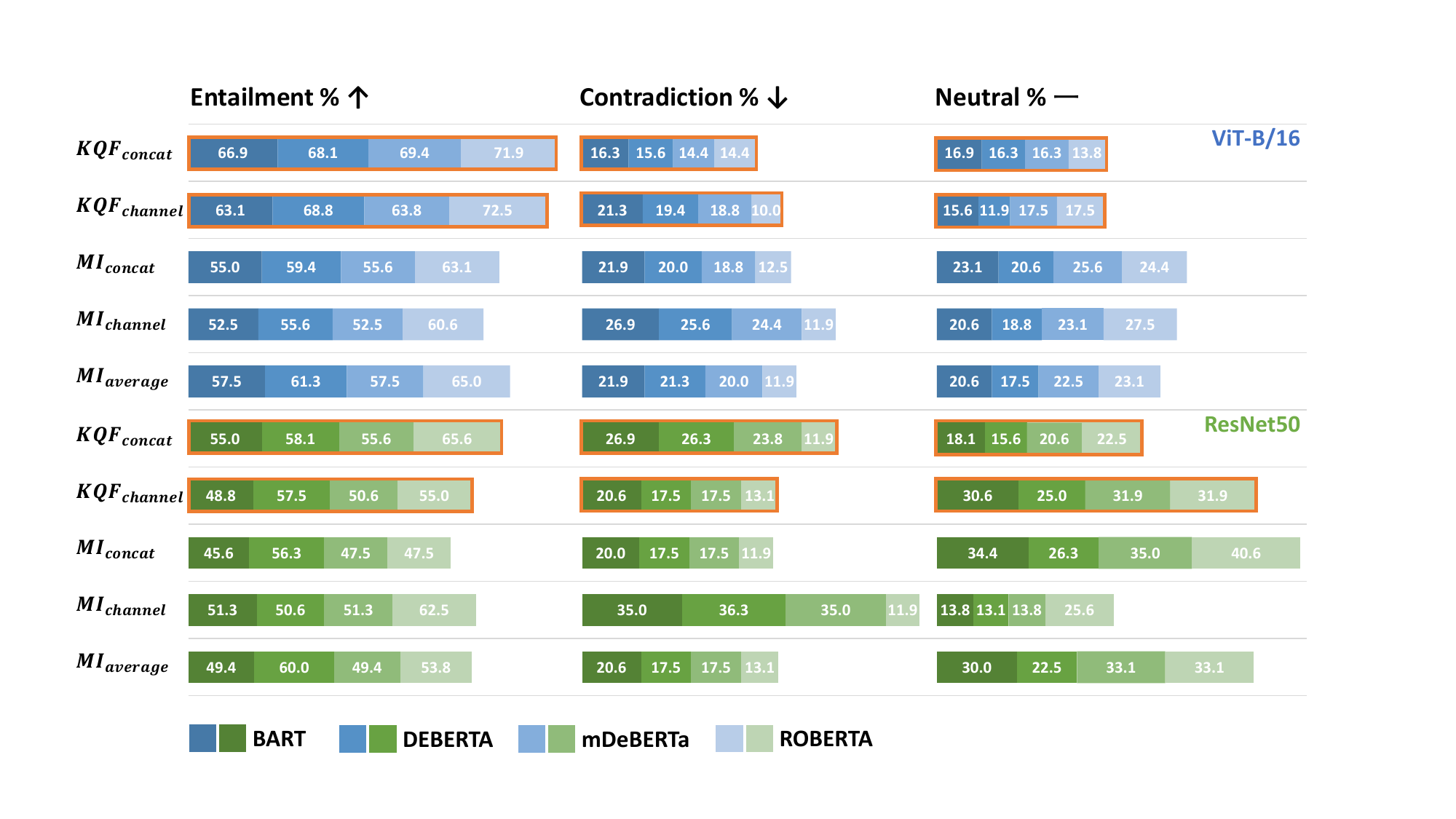}}
\caption{Evaluation results on open questions by different NLI models show that the \emph{KQ-Former} consistently achieves the highest entailment ratios and lower contradiction ratios compared to the baseline models across various tests.}
\label{fig::nli}
\end{figure*}

This section assesses the impact of language models on the explainability of anomaly maps. Figure~\ref{fig4} displays examples of the KQ-Former addressing radiologist queries. Instances (a, c, e) show the KQ-Former effectively describes anomaly maps. In cases (b, d, f) predictions do not fully align with expected outcomes. However, the language model still interprets these questions effectively and offers contextually relevant responses, demonstrating its capability to enhance the understanding of anomaly detection.

Table \ref{table1} summarizes the performance metrics. Independent of the backbone architecture or fusion strategy employed, the \emph{KQ-Former (KQF)} consistently outperforms the multi-image VQA baseline (MI) in all performance metrics. For instance, it improves accuracy by 5\% for closed questions and increases the BLEU-4 score by 71\% for open questions for the best variants. Among the different backbone architectures tested, the Vision Transformer (ViT) consistently outperforms the ResNet50. Specifically, switching to ViT boosts the KQ-Former's accuracy by 11.27\% for closed questions and improves its BLEU-4 score by up to 26.77\% for open questions. Regarding fusion strategies, the concatenation approach generally yields the highest improvements in both methods. We observe that \emph{KQF} is more robust across different fusion strategies, likely due to its enhanced ability to utilize visual features from multiple images effectively.
Additionally, the NLI model results depicted in Figure \ref{fig::nli} further validate the robustness of the \emph{KQ-Former}. The model demonstrates a higher entailment ratio and a lower contradiction ratio across different configurations, indicating that its answers are not only contextually appropriate but also more aligned with the factual content of the ground truth.

\subsection{Anomaly Maps Improve Generalizability of Language Models}
In this section, we investigate how anomaly maps enhance the generalizability of language models, particularly in detecting unknown anomalies. We utilized the top-performing KQF method with a ViT backbone for this experiment. Table~\ref{table2} shows that using anomaly maps with both concatenation and channel fusion strategies leads to better detection of known anomalies, with accuracy improvements of 3\% and 2\%, respectively. More significantly, anomaly maps greatly improve performance on previously unseen anomalies. For instance, including anomaly maps in the concatenation strategy raised overall accuracy from 69.67\% to 82.35\%, marking an 18\% improvement in identifying open-set anomalous data. These findings underscore the substantial role of anomaly maps in boosting the adaptability of language models.

\begin{table}[t!]
    \centering
    \setlength{\tabcolsep}{4pt}
    \caption{Performance in anomaly detection for known and unknown anomalies. The utilization of anomaly maps enhances performance in anomaly detection, particularly improving the VQA model's ability to generalize to previously unobserved pathologies.\label{tab::anomaly}}
    \begin{adjustbox}{width=0.95\linewidth,center} 
        \begin{tabular}{l  l | c c || c c | c c | cc}
            \toprule	    
         & \multirow{3}{*}{Method} & \multicolumn{2}{c||}{Known} & \multicolumn{6}{c}{Unknown} \\
        
           & & \multicolumn{2}{c||}{Overall} & \multicolumn{2}{c|}{Overall} & \multicolumn{2}{c|}{Unhealthy (17\%)} & \multicolumn{2}{c}{Healthy (83\%)} \\
           & & ACC $\uparrow$ & F1 $\uparrow$ & ACC $\uparrow$ & F1 $\uparrow$ & ACC $\uparrow$ & F1 $\uparrow$ & ACC $\uparrow$ & F1 $\uparrow$\\\midrule
           \multirow{2}{*}{\rotatebox{90}{Conc.}} & w/o Ano & 85.29 & 85.29 & 84.13 & 87.50 & 69.67 & 80.00 & \boldmath{$98.70$} & 95.00\\
           & w Ano. & \boldmath{$88.24$} & \boldmath{$88.19$} & \boldmath{$89.37$} & \boldmath{$89.37$} & \boldmath{$82.35$}\gtr{18\%} & \boldmath{$82.35$}\gtr{3\%} & 96.39 & \boldmath{$96.39$}\\\midrule 
           \multirow{2}{*}{\rotatebox{90}{Chan.}} & w/o Ano & 89.71 & 89.69 & 84.45 & 87.00 & 71.43 & 78.95 & 97.47 & 95.06\\
           & w Ano. & \boldmath{$91.18$} & \boldmath{$91.15$} & \boldmath{$85.72$}& \boldmath{$88.85$} & \boldmath{$72.73$}\gtr{2\%} & \boldmath{$82.05$}\gtr{4\%} & \boldmath{$98.72$} & \boldmath{$95.65$} \\
           \bottomrule
\label{table2}
\end{tabular}
    \end{adjustbox}
\end{table}

\section{Conclusion}

In this work, we integrated language models with unsupervised anomaly detection and introduced the first multi-image Visual Question Answering benchmark for anomaly detection (VQA-UAD). We established multi-image VQA baselines and analyzed various feature fusion strategies. We then proposed the \emph{Knowledge Querying Transformer (KQF)} module, which considerably enhanced the extraction of knowledge-related visual features when fine-tuned on a small dataset. Our findings demonstrated mutual benefits: language models provided interpretability to anomaly maps, improving clinical insights, while anomaly maps enhanced the generalizability of language models, particularly for detecting previously unseen anomalies.

Future work will explore larger language models trained on extensive medical knowledge and expand the diversity and size of our dataset. This will further enhance the generalizability and robustness of our anomaly detection framework across diverse healthcare settings. We believe our research will open new avenues for combining language models with unsupervised anomaly detection, driving innovations in this field.

\subsubsection{\ackname} C.I.B. is funded via the EVUK program (“Next-generation Al for Integrated Diagnostics”) of the Free State of Bavaria and partially supported by the Helmholtz Association under the joint research school ‘Munich School for Data Science’.

\bibliographystyle{splncs04}
\bibliography{MyLibrary}

\newpage
\title{Supplementary Material for Language Models Meet Anomaly Detection for Better Interpretability and Generalizability}

\author{Jun Li\inst{1,2} \and Su Hwan Kim\inst{4} \and  Philip Müller\inst{1} \and Lina Felsner\inst{1}  \and Daniel Rueckert\inst{1,2,4,5} \and Benedikt Wiestler\inst{1,4} \and Julia A.Schnabel*\inst{1,2,3,6}\textsuperscript{(\Letter)} \and Cosmin I. Bercea* 
 \inst{1,3}\textsuperscript{(\Letter)}} 

\authorrunning{Jun Li et al.}
\renewcommand{\thefootnote}{\fnsymbol{footnote}} 
\footnotetext[1]{Equal contribution.}

\institute{Technical University of Munich, Germany  \and
Munich Center for Machine Learning, Germany \and
Helmholtz AI and Helmholtz Munich, Germany \and
Klinikum Rechts der Isar, Munich, Germany \and
Imperial College London, UK \and
King’s College London, UK \\
\email{\{june.li,julia.schnabel,cosmin.bercea\}@tum.de}
}
\maketitle

\begin{figure*}[!h]
\centerline{\includegraphics[width=\textwidth]{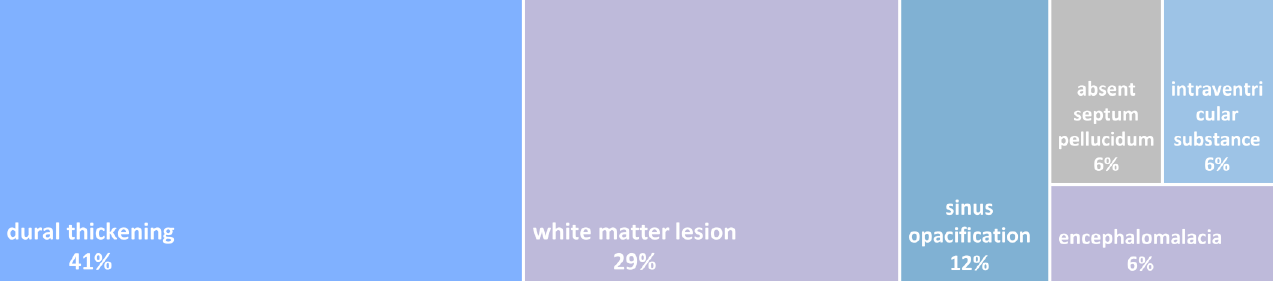}}
\caption{Category distribution of unseen anomalies. These unseen anomalies are dural thickening, white matter lesion, sinus opacification, encephalomalacia, intraventricular substance, and absent septum pellucidum.}
\label{fig2}
\end{figure*}

\begin{figure*}[]
\centerline{\includegraphics[width=\textwidth]{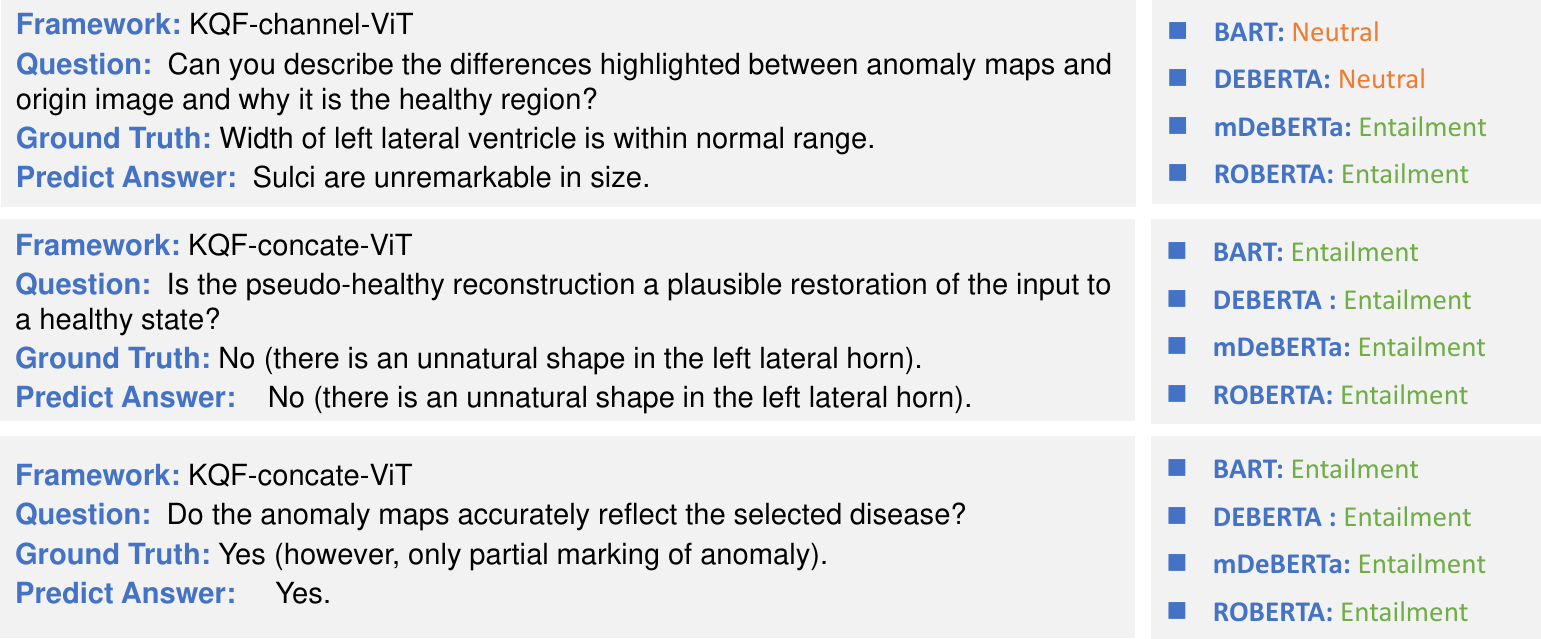}}
\caption{Visualization examples from different NLI models. In certain instances, different models may have different judgments, indicating that the results may still exhibit some deviations from human recognition. For example, in the first case, the KQF framework predicts \protect\say{Sulci are unremarkable in size} and the ground truth is \protect\say{Width of left lateral ventricle is within normal range}. The BART and DEBERTA models classify as \protect\say{Neutral}, while mDeBERTa and ROBERTA predict as \protect\say{Entailment}.}
\end{figure*}

\begin{table}[htb!]
\renewcommand{\arraystretch}{1.5}
\centering
\caption{Definitions and illustrative examples of questions and responses. Responses to open questions vary widely in format, so a single example is provided for visualization.}
\begin{adjustbox}{width=0.95\linewidth,center}
\begin{tabular}{ L{1.5cm} | L{5.5cm} | L{5.5cm} }
\hline
\textbf{Type} & \textbf{Question Definition} & \textbf{Response Examples} \\
\hline
\multirow{3}{*}{Closed} & Is the case normal? & Yes. / No. \\
\cline{2-3}
& Please describe the condition of the brain. & It’s + category. \\
\cline{2-3}
& Can you comment on the severity of the pathology? & Clinically irrelevant. / Potentially clinically relevant. / Clinically relevant. / Not applicable. \\
\hline
\hline
\multirow{4}{*}{Open} & Are there areas in the anomaly maps that highlight a normal variation of the healthy, rather than pathological areas (false positives)? & Yes. Anomalies are observed in the left frontal and right occipital sulci. \\
\cline{2-3}
& Is the pseudo-healthy reconstruction a plausible restoration of the input to a healthy state? & No (there is a midline shift to the right). \\
\cline{2-3}
& Do the anomaly maps accurately reflect the selected disease? & No (not marked). \\
\cline{2-3}
& Can you describe the differences highlighted between anomaly maps and the original image and why it is the healthy region? & Ventricles probably appear narrow because of the height of the slice. \\
\bottomrule
\end{tabular}
\end{adjustbox}
\end{table}

\end{document}